# PUPAE: Intuitive and Actionable Explanations for Time Series Anomalies

Audrey Der[1], Chin-Chia Michael Yeh[2], Yan Zheng[2], Junpeng Wang[2], Zhongfang Zhuang[2], Liang Wang[2], Wei Zhang[2], Eamonn Keogh[1]

[1]University of California, Riverside {ader003@ucr.edu, eamonn@cs.ucr.edu}
[2]Visa Research {miyeh, yazheng, junpenwa, zzhuang, liawang, wzhan}@visa.com

## Abstract

In recent years there has been significant progress in time series anomaly detection. However, after detecting an (perhaps tentative) anomaly, can we explain it? Such explanations would be useful to triage anomalies. For example, in an oil refinery, should we respond to an anomaly by dispatching a hydraulic engineer, or an intern to replace the battery on a sensor? There have been some parallel efforts to explain anomalies, however many proposed techniques produce explanations that are indirect, and often seem more complex than the anomaly they seek to explain. Our review of the literature/checklists/user-manuals used by frontline practitioners in various domains reveals an interesting near-universal commonality. Most practitioners discuss, explain and report anomalies in the following format: The anomaly would be like normal data A, if not for the corruption B. The reader will appreciate that is a type of counterfactual explanation. In this work we introduce a domain agnostic counterfactual explanation technique to produce explanations for time series anomalies. As we will show, our method can produce both visual and text-based explanations that are objectively correct, intuitive and in many circumstances, directly actionable.

## 1 Introduction

Recent years have seen significant advancements in Time Series Anomaly Detection, or TSAD, with state-of-the-art algorithms now outperforming even domain experts [24]. For example, consider the one-minute snippet of Electrocardiogram (ECG) data shown in Figure 1.

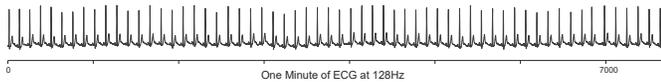

Figure 1: One minute of ECG data. From out-of-band data, we know that it contains a single anomaly, which must be very subtle.

Even with careful visual inspection, is not apparent to the human eye that this time series has any anomalies. Nevertheless, as Figure 2 shows, an anomaly *can* be discovered in the data.

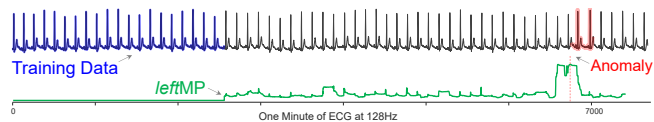

Figure 2: The ECG shown in Figure 2 processed with the *left*MP [48] set to find two-second long anomalies. The algorithm indicates an anomaly (highlighted in red) at location 6,763 (red dashed line).

Without the out-of-band confirmation that the algorithm did indeed find a true anomaly, it is not clear that we would trust this result. Even if we did trust it, perhaps recalling that this algorithm had performed very well on similar data, we would surely be curious as to *why* the algorithm had flagged this.

To this end, we conducted both interviews with domain experts and an extensive review of their literature on anomalous findings. We noted that domain experts often explain anomalies by noting what would have to change to make the anomaly appear normal, essentially completing the following template:

`Would be like A, except for corruption B`

Let us attempt this for the anomaly we discovered in Figure 2. Under the Euclidean distance (ED) that *left*MP algorithm used to detect this anomaly, it was 15.3 units from its nearest neighbor in the training data. As shown in Figure 3, if we simply replaced ED with the dynamic time warping (DTW) distance, that number drops to just 2.3.

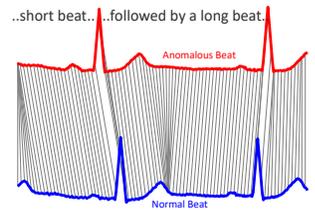

Figure 3: The anomaly compared to its DTW nearest neighbor in the training data.

This dramatic change suggests the cause of the anomaly was not a change in *shape* per se, but simply local changes in the *timing* of the beats.

As Figure 3 shows, from the computed warping alignment, it appears that the anomaly has a *short beat followed by a long beat*. In fact, if we Google that exact phase, the number 1 hit is a book that contains the text "*..compensation beats: **a short beat followed by a long beat**, which together last the same duration as two normal beats..*" [25]. According to noted cardiologist Dr. Gregory Mason, the anomaly we discovered is indeed a *compensation beat*, a particular type of supraventricular arrhythmia.

In Figure 4.*left* we show another anomaly discovered later in the same dataset.

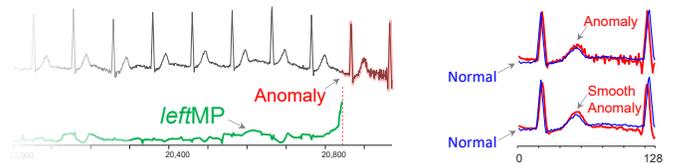

Figure 4: *left*) Another anomaly found in the ECG. *right*) The anomaly overlayed with its nearest neighbor before and after smoothing.

For this anomaly using the DTW distance instead of ED makes no difference. However, as Figure 4.*right* suggests, a little smoothing drastically reduces the distance between the anomaly and its nearest neighbor. This suggests that this anomaly is caused by, and therefore best explained by, *noise*.

These examples outline our basic approach to time series anomaly explanation. Given an (perhaps tentative) anomaly, we will test a small set of operators that may be able to transform the anomaly into a normal data pattern. The most parsimonious transformation (or rather, its *inverse*) can be seen as the best explanation of the anomaly. This idea opens two challenges:

- What is the correct set of operators that is expressive enough to explain all anomalies?
- Assuming that some of these operators are in different units, how do we make them commensurate, so that they can be ranked, and the best explanation offered?

In this work we introduce principled answers to both questions. We demonstrate the intuitiveness, and the correctness and actionability of our explanations on diverse domains.

The rest of this paper is organized as follows. Section II contains our motivation and Section III our closest related works. In Section IV, we introduce our methodology and the operators it leverages. In Section V, we discuss how we made them commensurate and comparable. In Section VI, we present our empirical results in the form of a benchmark.

## 2 Motivation

The recent successes in the community's ability to *find* anomalies have outpaced its ability to *explain* them, once discovered. It is not even clear what format an anomaly explanation should take. The literature is replete with ideas [13] [41]; weight matrices, heatmaps and feature-saliency maps [19][42], rule-based systems [10], weighted trees [8], etc. have all been employed.

However, many of these approaches seem very indirect. In some cases, the explanations seem to be more complex than the data they seek to explain. In addition, for most of these techniques, the method of anomaly *explanation* in inextricability tied to the method of anomaly *discovery*. As we will later argue, it is advantageous to divorce these two steps.

Returning to our original question, what format should an anomaly explanation take? Rather than impose our preferred method on the users of anomaly detection systems, it makes sense to understand how domain experts normally talk, reason and communicate about anomalies. The follow examples give a flavor of this (emphasis ours):

- "*..similar to pattern 1 **except** for a deep valley*" [12].
- "*..like ATMs 1 and 2, **except** for a large spike*" [43].
- "*..similar to the vertical spectra **except** for the peak*" [22].
- "*..similar to western SIO **except** it shows two crests*" [29].
- "*This is similar to inhibition-induced spiking, **except** that the response is a burst.*" [23].

Note that while these examples are from diverse domains, they all have the same basic structure. In each case the explanation says *what* corruption must be removed (equivalently, what edit could be performed) in order to make the anomaly look like *what* normal pattern.

For example, the authors of [12] suggest that if the anomaly is edited to remove the deep valley, it would look like a normal pattern 1. Such explanations are often written in the form of:

```
Would be like A, except for corruption B
```

The reader will note that such explanations are *counterfactual explanations*. As our examples suggest, such explanations appear to be universally used in industry [3][40], science and medicine.

*Counterfactual examples* (or *counterfactuals*) are a popular form for explanations in explainable AI (XAI), perhaps because they are amenable to the way people "*select to mutate in their representation of reality*" [11]. Counterfactuals (CFs) describe what should be different in a system's input to produce a different outcome, allow for imagining better (as opposed to worse) outcomes, and do not focus on improbable events [11]. CFs are thought to be particularly helpful if they are *sparse* and *proximate* [21].

These observations motivate PUPAE (Personalizable Universal Plausible Anomaly Explanations), our proposed definition of anomaly explanation. The formal definition is in Section 5.1, but can be informally stated as: A PUPAE is the minimum change that can be made to an anomaly to make it similar to some normal data at location *Loc* in the training data.

Note the location *Loc* in the training data is an intrinsic part of the explanation. For example, consider these two following explanations:

- The pattern looks like Monday Dec 25$^{th}$ 2023 if we remove the spike at the beginning of the anomaly.
- The pattern looks like Monday Dec 4$^{th}$ 2023 if we remove the spike at the beginning of the anomaly.

Note the distortion is the same in both cases, but in the former case, we can see that this anomaly is more similar to a holiday, than a regular working day.

## 3 Related Work

Several works explore the use of natural language counterfactuals for various domains, including NLP [32] and computer vision [18]. However only a handful of research efforts have considered counterfactual explanations for *time series*. Additionally, out of the methods we discuss in this section, PUPAE is the only method that uses *natural language counterfactual explanations for time series*. Some methods to explain and visualize time series tasks include by way of clustering [39], feature importance and heatmaps [19][42], or in terms of other subsequences of the time series accompanied by natural language [40]. TsXplain [28] generates a paragraph of textual based on results from a statistical feature extractor, listing numerical values for statistical features.

PUPAE can be best be categorized as a *perturbation-based method*, which directly computes the contribution of the input features by altering them, computing over the altered input, and measuring the difference between the original input [1]. Other perturbation-based methods include *Native Guide* [15] and TeRCe [4]. However, these methods generate synthetic time series subsequences to visually juxtapose with the anomaly; In contrast PUPAE uses only real data and generates *natural language* counterfactuals.

## 4 Counterfactual Operators

We are now able to explain our proposed operators. These operators are based on our inspection of all publicly available TSAD benchmarks [17][24][44][45], reviewing the literature [2][12][22][25][36], and interviews with several domain experts in cardiology (Dr Gerg Mason), oil&gas processing and entomology (Dr. Kerry Mauck). While we shall show that these operators are very expressive, we do not claim that they are complete. However, our framework is general enough that it would be easy for a practitioner to augment our system with a custom domain-specific operator.

### 4.1 A Motivating Domain in Entomology.

To ground our examples in reality, we searched for a single domain that could illustrate our proposed operators. Entomology offers such a possibility. Phytophagous (plant eating) insects such as the Asian citrus psyllid (*Diaphorina citri*) can have their behavior monitored by a device called an electrical Penetration Graph (EPG). Datasets collected this way are routinely searched for anomalies, which may be novel insect behaviors, or data artifacts that must be excluded from analysis.

Normally we envision the training data for a TSAD algorithm as being a single long time from which subsequences are sampled [24][44][45]. However, for simplicity, in Figure 5 we show just a curated (by entomologist, Dr. <blinded>) set of subsequences.

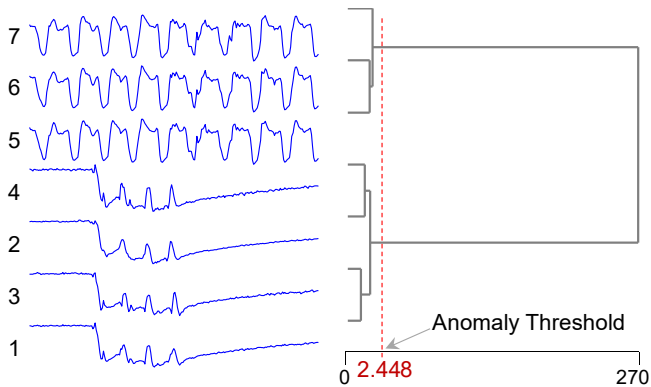

Figure 5: A set of subsequences which act as a training dataset for our simple TSAD model. Note that the dataset [46] is polymorphic, there are two types of normal behaviors, known to entomologists as G (1 to 4) and E2 (5 to 7).

In our training data, the mean z-normalized Euclidean distance between an instance and its nearest neighbor is 2.105, with a standard deviation of 0.114. Thus, we can define as anomalous, any time series that is not within μ +3σ (2.448) of at least one subsequence. This is the *anomaly threshold* shown in Figure 5, just below the dendrogram.

This may seem like a trivial TSAD algorithm, but it is essentially equivalent to the Matrix Profile algorithm, which is considered among the state of the art (SOTA) for time series anomaly detection [24]. In any case, recall that our aims here are completely independent of the TSAD algorithm used.

Below we discuss the set of operators in our framework. For brevity we mostly discuss them at a conceptual level. However, in every case, we have both detailed pseudocode and actual code available at [49].

### 4.2 Operator: Uniform Scaling.

The first anomaly we consider is 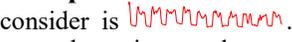. It is not obvious why this is an anomaly, as it strongly resembles patterns 5 to 7 in the training set. However, it is 24.57 from its nearest neighbor in the training data, which is much larger than the anomaly threshold.

In Figure 6.*left* we show the effect of taking this anomaly, "stretching" it in the time axis, and (re)comparing it to pattern 7. In order to make the stretched versions of the anomaly commensurate with the fixed length training data, we must truncate off the surplus datapoints (highlighted in yellow).

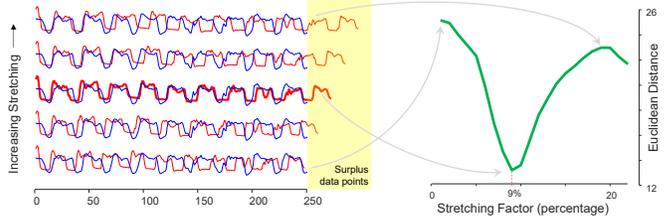

Figure 6: *left*) The anomaly (red) is far from its nearest neighbor (blue) because it is faster (has higher periodicity) that its nearest neighbor. By "stretching" it and recomputing the distance, we find that it closely matches normal data if it is nine percent longer (*right*).

In Figure 6.*right* we show the effect that the rescaling has on the Euclidean distance. As we increase the stretching, the distance begins to drop dramatically, minimizing when the anomaly is 109% of its original length, and then rising sharply again. We can interpret this as an explanation in our proposed format:

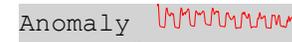

In Figure 6 we only consider *stretching* the anomaly, however we also *shrink* the anomaly, by reversing the roles in the computation. There is an important caveat to this operator: we must place some limits on the amount of rescaling. When any time series stretched arbitrarily long, it effectively becomes a straight line, and straight lines have a surprisingly low distance to any time series [6]. Without loss of generality, we limit the rescaling to between 80% to 120% of the original length.

### 4.3 Operator: Occlusion.

Anomalies like this 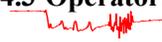 are very common in many domains. The overall shape is familiar, but there is a *local* burst of noise or other distortion. This has a natural explanation in our framework, the anomaly would be like a normal object if we ignored a sub-region of some *length*, starting at some *location*.

This means that we need to have some technique to predict the appropriate sub-region to occlude. It is clear we must charge some penalty for any data we ignore, otherwise we can always reduce the distance between any pair of time series by ignoring some data.

As shown in Figure 7 we can conduct a grid search over all possible values for *location* and *length*, scoring each combination. For our sample dataset, there is a strong peak (the darkest pixels) at *location* =157 and length = 37, which seems to be the correct answer.

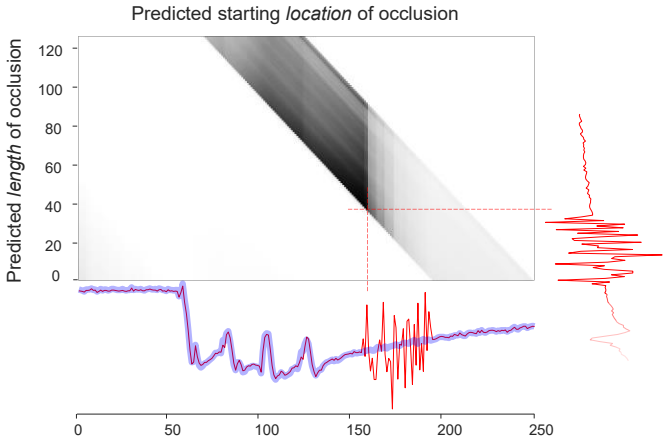

Figure 7: The algorithm outlined in Table 1 creates a heat map whose darkest value predicts the length and location of the subsequence that we should ignore to maximize its similarity to a normal data object.

We now can interpret this in our proposed format:

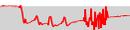

Anomaly `⎯⎯` would be like **3**, **except** for **region of length 37, from 157**

We can further expand our explanation, by noting that there are three common cases for the occluded data. If the mean of the occluded data is about the same as the mean of the remaining data (as in the above case), we can echo `noise`, if the mean is at least one standard deviation greater than the mean of the remaining data, we can echo `spike`, and if the mean is at least one standard deviation less than the mean of the remaining data, we can echo `dropout`.

In Table 1 we formalize the algorithm that conducts the grid search to find the best occlusion explanation.

Table 1: The Occlusion Euclidean distance (OED).

```
Algorithm: Occlusion
Input: T (model series), A (anomaly series)
Outputs: Location, Length, OED (Occlusion distance)
1  L ← length(T) % where A is of equal length
2  dists ← Inf(L,L/2)
3  OCCL_SCALE ← linspace(0,2,L/2)
4  for olen=1:L/2 do              % test occlusion lengths
5    for oloc=1:L-olen do          % test starting index
6      if olen+oloc > L then break
7      b,a ← oloc-1, oloc+olen
8      before ← ||T(1:b)-A(1:b)||₂
9      after ← ||T(a:end)-A(a:end)||₂  % mean of elem-wise diff
10     O1 ← abs(mean(T(1:b)-A(1:b)))
11     O2 ← abs(mean(T(a:end)-A(a:end)))
12     O ← ((O1+O2)*0.5)*olen
13     dists(oloc,olen) ← before + after + O + OCCL_SCALE(olen)
14   end for
15 end for
16 return min(dists(:))
```

Lines 1-3 initialize relevant constants and the data structures that the results will be stored in. Then in Lines 4-16 we compute and store the Occlusion Euclidean distance (OED) in the 2D distance matrix. For any occlusion length and occlusion location, the OED is computed in lines 7-13. In lines 7-9, The ED distance between the model series *T* and anomaly series *A* for the subsequences before and after the occluded subsequence. The core logic of the operator is in lines 10-12. Framing this as a counterfactual, we ask, "*If the anomalous region were to instead resemble a normal signal instead, what would the distance realistically be*?" Following this line of thought, we substitute the distance of each time step in the anomalous region with the average distance between the model and anomaly series where the signal indicates typical behavior. The penalty term based on `OCCL_SCALE` (line 13) is to prevent the algorithm from choosing to occlude as much of the series as possible.

There is another more general thing we can do to enhance the intuitiveness of our explanations. In some cases, instead of reporting the *relative* location of the anomalous subregions, we can report the *absolute* time. This may give the user additional insights. For example, suppose that a UK electrical grid operator noticed an anomalous `spike` at 3:45pm on 18/12/2022. She may search her mind for an explanation "*what could have caused this? Ahh it was half time in the World Cup final!*" (It has long been known that in England there is a large spike in electrical power demand during the half time of games, as millions of households simultaneously turn on their kettles [38]).

To illustrate both ideas, we applied the DAMP algorithm [24] to EnerNOC887, a year-long energy usage dataset for an industrial site [2]. DAMP easily found the true anomalies, and as Figure 8 shows, two of them happen to be in the same month.

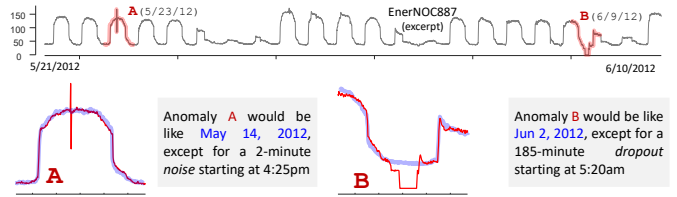

Figure 8: *top*) A three-week long snippet of EnerNOC887 has two anomalies. DAMP found them, and PUPAE (*bottom*) explains them. Note that **A** is reported as `noise`, as the `spike` is followed by a `dropout` of about the same magnitude, giving the offending region a mean value that is similar to the original data.

**4.4 Operator: Warping.** Consider this anomaly 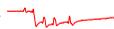, once again is not obvious why this is an anomaly, as it appears to resemble patterns 1 to 4 in the training set. However, if we overlay it with its nearest neighbor in the training data (exemplar 3, shown in blue), we can see that some of the peaks are out of phase 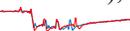.

The reader will appreciate that this misalignment of features is called informally called "warping", and dynamic time warping (DTW) is a distance measure that is designed to be invariant to this. In particular, the anomaly has a Euclidean distance of 4.2 to its nearest neighbor in the training set, and this is reduced to 2.3 under the DTW distance.

Note that for any two time series, their DTW distance is always less than or equal to their Euclidean distance. Thus, just as with the Occlusion and OED operators, a penalty distance needs to be incorporated into the DTW distance if it is to be compared to the ED of the other operators.

Where *A* is the anomaly subsequence, $T_N$ is its nearest anomaly-free nearest neighbor, $L_A$ the length of the anomaly, $A'$

the warping path, and $L_{A'}$ the length of the warping path, compute and set $I_{warping}$ as follows:

$$I_{warping} = \frac{DTW(A,T_N)*(L_A/|L_{A'}-L_A|)}{ED(A,T_N)}$$

As discussed in more detail below, a greater reduction in distance relative to the Euclidean distance between the anomaly and its nearest neighbor will semantically indicate that the operator is a plausible explanation for the anomaly. Scaling $DTW(A,T_N)$ according to the reciprocal of difference in length according to the warping path allows PUPAE to communicate that anomalies fully utilizing the DTW warping window are more likely to be appropriately explained as a *warped* anomaly.

While we recognize this is subject to an edge case when the lengths of $L_A$ and $L_{A'}$ are of equal length, but the reader will appreciate that this particular edge case of a warping path of length zero is semantically Euclidean distance, and this is easy accounted for before computing $I_{warping}$.

### 4.5 Operator: Smoothing.

Consider this anomaly 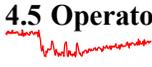, it is easy to see that it resembles patterns 1 to 4 in the training set but is simply *globally* noisy (as opposed to the example in Figure 7 which had *local* noise). The anomaly has a Euclidean distance of 3.0 to its nearest neighbor in the training set, and this is reduced to 2.2 after it is smoothed using the simple moving mean algorithm with a window length of three.

We have found that such anomalies are important to be able to explain, as their meaning is highly domain dependent and actionable. For example, in many medical contexts such noise is virtually always due to "Power Line Interference" [5], and uninteresting to the clinicians. In fact, it is often cited as the number one source of *false positive fatigue* [37]. In contrast, in some oil&gas operations, short bursts of high frequency noise can be caused by cavitation due to large pressure drops. Such anomalies need to be prioritized for investigation, as they are often precursors to dramatic failures. While there is significant literature on smoothing algorithms, we have found a simple moving average filter seems to be universally useful here.

### 4.6 Operator: Reversal (LR Flip).

We synthetically created this anomaly 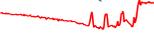 to remain consistent with our insect dataset. However, these reversed shapes *do* appear as anomalies in the wild [47]. The anomaly has a Euclidean distance of 17.6 to its nearest neighbor in the training set, and this is dramatically reduced to 0.54 after it is flipped horizontally.

### 4.7 Operator: Flip Upside Down (UD Flip).

This example 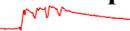 is also a synthetically created anomaly to remain consistent with our insect theme. However, once again these Y-axis reversed shapes *do* appear in the wild as anomalies [47]. The anomaly has a Euclidean distance of 23.27 to its nearest neighbor in the training set, and this is reduced to 1.95 after it is flipped vertically.

### 4.8 Operator: Linear Trend.

Consider this anomaly 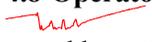. As with some of the previous anomalies, it resembles patterns 1 through 4 in the training set. However, with careful visual inspection the reader may realize that it has a slightly greater upward linear trend.

There are standard algorithms for detrending time series, which are logically equivalent to *finding* the best fitting line to the data, then *subtracting* that line from the data. Before we apply this correction to the above, the distance to its nearest neighbor in the training set is 8.7, afterwards the distance is reduced to just 0.93.

This is another example of an anomaly explanation with clear actionability. In patient monitoring in an ICU setting, a change in linear trend of a *PPG signal* (oxygen saturation) is usually ignorable. It is typically due to patient movement. However, a change in the linear trend of a *respiration signal* may indicate dyspnea (difficulty breathing) caused by upper airway obstruction and should be attended to immediately.

### 4.9 Operator: Piecewise Normalization.

Consider this anomaly 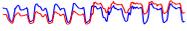, which we show overlayed with a normal time series (in blue) for context. The reader will appreciate the two time series look very much alike, except that the anomaly has a "level-change" or "shift" in its middle. Such anomalies seem to be very common across multiple domains. The change can be a mean change (as in the above, and/or an amplitude change.

The cause of such changes is domain dependent. In our insect EPG example, there is a situation that is known to cause this anomaly. The insect being monitored has a fine uninsulated gold wire (about 1/100$^{th}$ the thickness of a human hair) glued to it and the insect is allowed to roam around on, and feed on a plant. If that wire should touch a part of the plant, it reduces the electrical resistance, and causes a level shift. In the example shown above, the change point occurs in the middle of the time series, however we cannot be sure that this is generally true, especially as we wish to be TSAD algorithm independent. Thus, we will need to test all possible split points.

This anomaly can be corrected by piecewise normalization. In particular, for two time series $A$, and $B$, both of length $m$, we can compute the Piecewise Normalized Distance:

$$PND(A,B) = \underset{1 \leq i \leq m-3}{argmin}\left[\sqrt[2]{ssd(A_{1:i},B_{1:i}) + ssd(A_{i+1:m},B_{i+1:m})}\right]$$

Where *ssd* is the sum of squared distances between two z-normalized vectors: $ssd(X,Y) = \sum_{j=1}^{|X|}(X_j - Y_j)^2$

The *PND* reduces to the ED if the left and right sides of both time series have the same means and standard deviations but produces a dramatically lower score in the presence of a shift of mean and/or standard deviation. In Figure 9 we show an example of PND applied to a section from a classic benchmark.

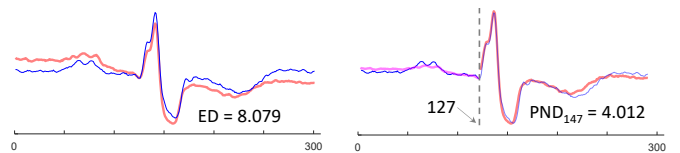

Figure 9: *left*) An apparent anomaly from the MITDB [26] benchmark overlaid on its nearest neighbor. *right*) The PND operator suggests

cutting the anomaly at 127 to produce a dramatically reduced distance, and to visually reveal the latent similarity.

The time series in Figure 9 is *not* the labeled anomaly in this dataset, but is flagged as an anomaly by some algorithms including the Matrix Profile and Telemanom [20] (on *some* runs; Telemanom is stochastic). This hints at the value of time series explanations. If a user is experiencing many false positives, but they all have the same explanation, she can suppress the reports that have that explanation, or refine her algorithm.

**4.10  Discussion of Operators.** As we will show in our experimental evaluation section, the list of operators above is very expressive, and covers the vast majority of anomalies available in the communities' benchmark datasets [44][45]. However, we emphasize that we are not claiming that these are a *complete* set. We believe that our framework is general enough to allow practitioners to add custom operators specialized for their domain.

PUPAE only allows a single operator to be an explanation. The reader may argue that an anomaly may have two or more causes, say a spike *and* some warping. Again, based on the standard benchmark datasets [44][45] and our discussion with domain experts, this seems very rare, thus we defer further discussion of this to future work.

In addition to adding new bespoke operators, occasionally a user may wish to *remove* operators based on their domain knowledge. For example, they may know that their domain has a lot of inconsequential warping, thus warping should not be used to explain an anomaly.

## 5  Making Operators Commensurate

In the previous section we listed the full set of operators we consider in this work. However, we need to make all the operators commensurate so that we can compare and rank them, and only report the most plausible explanation to the user.

All our proposed operators return distances that are either ED or based on ED (note that DTW is simply the Euclidean distance on the "dewarped" time series). However, this does not mean that they are commensurate. For example, the occlusion operator may be reporting a distance based on only 50% of the data, whereas the reversal operator considers all the data.

For all operators we define a scalar $I$ (Improvement), which is simply the ratio of the new, post-operator distance divided by the original distance. For example, for an anomaly $A$, and its nearest neighbor $T_N$ in the training data we define:

$$I_{LinearTrend} = \frac{ED(A', T_N)}{ED(A, T_N)}$$

Where, in this example, $A'$ is the transformed anomaly parameterized by the amount of linear trend to add to $A$ such that $ED(A'_{LinearTrend}, T_N)$ is minimized. The other operators are defined similarly, and specifics are noted in Section 4. Note that there is no possibility of a division-by-zero error, as the operators are only used on time series that have a distance to their nearest neighbor that greater than some positive anomaly threshold. We note that $I$ score values can be greater than one, for example, $I_{LRFlip}$ can be much greater than one.

When selecting an explanation, the reader will appreciate that the method is rather simple and interpretable. For each operator, compute the $ED(A', T_N)$ for all combinations of values its parameters might take on. PUPAE reports the transformation (and any parameters) that minimize the $ED(A', T_N)$, as a lower $I$ score indicates an operator's success. This is formalized in Section 5.1. While the idea of conducting a search may seem computationally daunting at first glimpse, as aforementioned, only *Occlusion* has more than one parameter. Additionally, due to the semantic nature of these operators, the search space of each parameterized operator is further restricted.

For example, Piecewise Normalization communicates that, primarily, normalizing two subsequences piecewise will yield the greatest reduction in ED whilst being the most systematic change. This indicates that if PUPAE's best attempt at using Piecewise Norm to explain an anomaly results in normalizing a subsequence of small length separately, piecewise normalization is unlikely to be the most proximate explanation.

**5.1 A formal definition for PUPAE.** With our operators defined and the intuitive understanding of Improvement scores discussed, we can formally define the anomaly explanation that PUPAE provides (Equation 1). Given an anomaly-free time series $T_N$, an anomaly time series $A$, where $|T_N| = |A|$, and a set of operators $\mathcal{F} = \{f_0(\theta), f_1(\theta), \ldots, f_n(\theta)\}$:

Equation 1: Anomaly Explanations provided by PUPAE.

$$f_\theta^* = \underset{f_{i,\theta} \in \mathcal{F}}{\mathrm{argmin}} \left( \min_\theta \frac{ED(\hat{A}, T_N)}{ED(A, T_N)} \right), \hat{A} = f_\theta(A)$$

The input and output are mappable to the natural language counterfactual explanation template in Section 1:

`A` would be like `T`<sub>`N`</sub>`, **except** for corruption -f*`<sub>`θ`</sub>

Where $f_\theta^*$ is an optimal transformation, $-f_\theta^*$ is the perturbation for which it reverses. In the cases of parameter-free operators (e.g. LR Flip, UD Flip), there is no sign.

**5.2 Selecting Operators to include in $\mathcal{F}$.** While we assert that the operators discussed in this section are a comprehensive base set of operators, the operators in $\mathcal{F}$ should be selected with domain expertise. For example, in medical telemetry, changes in the linear trend may be considered a "wandering baseline" and are medically irrelevant. However, in industrial processes, a change in linear trend can be significant, and may indicate a batch process is "running hot". However, for simplicity, and to stress test our ideas, we do not do any domains customization in our experiments.

## 6  Empirical Evaluation

To ensure that our experiments are reproducible, we have built a website [49] which contains all data and code for the results. We have documented concrete details of our experiments to allow for reproduction of all our experiments. All experiments were conducted single-threaded on an Intel® Core i7-10710U CPU at 1.10GHz with 16 GB of memory.

We provide two types of empirical evaluation.

- Case studies that are designed to test our method on real world datasets. By nature, these studies are mostly anecdotal. However, in every case we work with a domain expert to help interpret our findings.
- Large scale experiments on corrupted datasets. Here, to avoid the conflict of interest of creating new corruption models to test our ideas, we work only with *existing* corruption models specially designed for TSAD.

The time needed to compute all our operators is inconsequential compared to the time needed for the original anomaly detection [24]. Thus, our evaluation focuses on the following questions: where objective ground truth exists, does PUPAE correctly discover it? And, in the absence of ground truth, is the explanation plausible to a domain expert?

Without loss of generality, expect where otherwise stated, we use the *left*Matrix Profile to discover the anomalies [24]. However, we remind the reader that PUPAE is completely independent of the algorithm used to do the actual TSAD.

### 6.1 Testing Individual Operators.
Before testing and demonstrating the PUPAE framework, we will begin by testing some of the operators in isolation.

The occlusion operator needs to predict the *location* and *length* of a subregion to delete from a time series to make match the ground truth of a distortion (noise, spike, dropout) that some anomalous process added to otherwise normal data. In the absence of large amounts of data for which the ground truth is known, we created synthetic examples by corrupting exemplars from the UCR archive [14].

We randomly select an exemplar *A*, from a UCR dataset, we then randomly select a value for *location* from 1 to ½ *m*, and a value for *length* from 0 to ¼ *m*. Note that the length of the distortion could be zero, and we would hope that our algorithm predicts that. We then choose randomly, with equal probably from the following three distortions: Add *noise*, add a *spike*, add a *dropout*. Figure 10 shows some examples in the wine dataset.

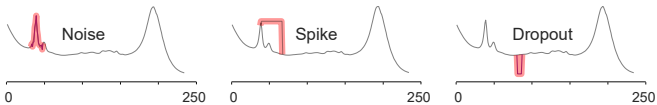

Figure 10: An exemplar from the UCR wine dataset, with three types of synthetic anomalies (highlighted in red) added.

Can we correctly recover the location and length of these distortions? We find the exemplar's (non-self) nearest neighbor *N* under Euclidean distance in the same dataset and use the operator in Section IV.C to predict the *location* and *length*. Figure 11 shows the results of 1,000 such tests.

The histogram bin sizes in Figure 11 are set to be as granular as a single datapoint to best communicate the precision and accuracy of the Occlusion operator's success at locating the location and length of the anomaly. Note that in almost all cases, PUPAE is able to correctly locate the anomaly (spike, dropout, or noisy subsequence) location and length. Perfect results are probably not possible, due to natural anomalies within the data.

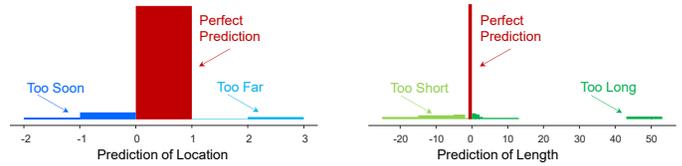

Figure 11: A benchmark of predictions created from the wine dataset [14], where one of three anomalies of random lengths are inserted into exemplars at random locations. From there, the Occlusion operator is run on each anomaly series and the source dataset, excluding the donor series the anomaly was inserted in. PUPAE selects the nearest neighbor to compare the anomaly to, computing a location and length of an occlusion to apply.

### 6.2 Multiple Short Case Studies.
To show the generality of PUPAE, here we will show some brief examples on diverse real datasets. In every case, the reader can find detailed provenance at [49].

**The power demand dataset** from UCI [16] has several annotated anomalies. As Figure 12 shows, two of them happen to be in the same week and can be found by the Matrix Profile.

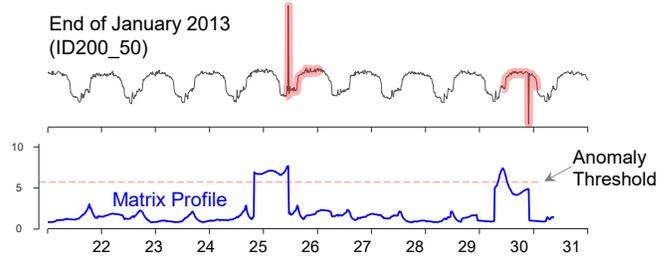

Figure 12: A snippet of a four-year electrical demand dataset happens to have two (easy to find) anomalies in the same week.

PUPAE described these as:

- `Anomaly` ⌐⌐ `would be like` **`23/01/2013`**, **`except`** `for` **`Spike at 4:30am`**
- `Anomaly` ⌐⌐ `would be like` **`29/01/2013`**, **`except`** `for` **`Dropout at 7:45pm`**

Here PUPAE's distinctions between "dropout" and "spike" may be actionable. A line engineer may prioritize the investigation of spikes (can be caused by lightning strikes or upstream tripped circuit breakers) over dropouts (typically caused by a faulty sensor, not a real change in power demand).

**The accelerometer dataset** is a dataset measuring the health of a cooling fan [17]. By manipulating magnets around the fan (whose blade was magnetic), the creators were able to induce various types of anomalies. The original authors performed some data transformations and created a sophisticated domain-informed neural network then discover these anomalies, however, as shown in Figure 13, we found that a direct application of DAMP on original data could find the anomalies.

The cause of the anomaly is not obvious, even with a careful visual inspection of the data, however PUPAE correctly states:

- `Anomaly` ∿∿∿ `would be like` **`data at 70.2 seconds`**, **`except`** `for` **`-9.1% uniform scaling`**

The anomaly reflects a sudden braking effect to the motor.

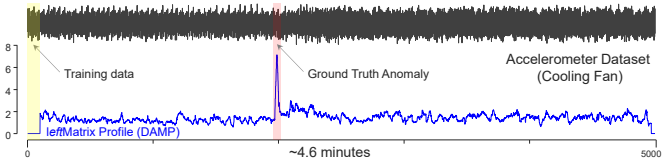

Figure 13: *top*) An five-minute snippet from the accelerometer dataset. *bottom*) DAMP correctly finds the anomaly, but what was its cause?

**6.3 Amazon's Benchmark.** A recently published paper by a research group in Amazon [17] includes a sophisticated tool for generating plausible synthetic anomalies. In Figure 14 we show *their* plot introducing this tool.

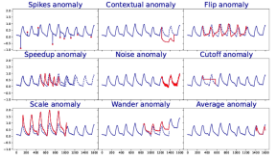

Figure 14: A screen capture of a figure from [17], showing examples of nine types of synthetic anomalies. The reader will appreciate that almost all the listed anomalies have counterparts in our operator set.

The authors argue (in a similar vein to our discussion of which operators to use in PUPAE) that the majority real-world anomalies are well modeled by these nine cases. The reader will appreciate that this model offers the opportunity to evaluate our ideas. We propose to corrupt anomaly-free data using a random choice of one of the above corruption models, and then test our ability to first *discover*, then *explain* the anomaly.

While Amazon's vision mostly overlaps with ours, before proceeding, we need to do a little (re)mapping between their terminology/model and our own, and slightly extend Amazon's model to include additional distortions. In particular:

- We limit Amazon's *spikes* to a single *spike*.
- Warped anomalies are added as an extra corruption.
- Flip Upside Down (UD Flip) are added as a corruption.

We omit a few of Amazon's distortions:

- Both *Contextual* and *Average* anomalies are polymorphic, and one of several different operators might be a solution for any instance, confounding our evaluation metric.
- *Cutoff* anomalies may be solved by an occlusion (as demonstrated in Section 6.1). Moreover, they are *trivial* anomalies in the sense, noted by Wu & Keogh [45].

Finally, to clarify differences in our naming conventions:

- Amazon's *Flip* anomalies are our *LR Flip* anomalies.
- Amazon's *Speedup* maps to our *Uniform Scaling* anomalies.
- Amazon's *Noise* maps to our *Noisy* anomalies.
- Amazon's *Scale* maps to our *Step* anomalies.
- Amazon's *Wander* maps to our *Linear Trend* anomalies.

We can now describe our experiment in detail. We:

- Randomly select a 10,000 datapoint long anomaly-free time series from a set of four: PPG, Gait, electrical demand fingerprinting [34] or conveyor belt system. To the best of our knowledge, after consulting with domain experts and examining out-of-band data, these datasets are anomaly-free.
- With uniform probability, randomly select one of the nine anomaly types, and insert an anomaly of this type at a random location between 3,001 and 9,000. The first 3,000 datapoints are reserved for training data, and we stop at 9,000 to avoid "edge-effects" with our sliding window.
- Using DAMP [24], with the first 3,000 datapoints as training data, attempt to find the anomaly. If the anomaly was successfully discovered, use MASS [31] to locate the anomaly's nearest neighbor in the anomaly-free training data.
- Use PUPAE to predict the corruption producing the anomaly.

Our benchmark has 300 exemplars of each anomaly type per dataset, totaling 2400 anomalies sourced from each dataset, with a benchmark size of 9,600 time series. Table 2 summarizes the results of PUPAE's ability to correctly identify the anomaly inserted, based on the benchmark formulation described above. Of the 9,600 trials, DAMP successfully discovered the anomaly 7,831 times. Confusion matrices separating this benchmark by dataset are available on our companion website at [49].

Table 2: A confusion matrix in which the rows indicate the type of anomaly inserted, and the columns indicate PUPAE's explanation. Where there are no values listed in a cell of the matrix, there were no missclassifications of that type made.

|  | LR Flip | LT | U Rescale | Noisy | UD Flip | Step | Warp | Spike |
|---|---|---|---|---|---|---|---|---|
| LR Flip | 1198 |  |  |  |  |  |  |  |
| Lin'Trend |  | 985 | 7 |  |  | 4 |  |  |
| U Rescale |  |  | 1200 |  |  |  |  |  |
| Noisy |  |  |  | 1200 |  |  |  |  |
| UD Flip |  |  | 1 | 24 | 662 |  |  |  |
| Step | 25 | 230 | 182 | 3 | 26 | 671 | 5 | 31 |
| Warp |  | 38 | 286 | 164 |  |  | 396 | 23 |
| Spike |  |  | 16 | 156 |  |  |  | 298 |

The results show PUPAE correctly explains the inserted anomalies with a mean per-dataset accuracy of 84.25%, and a general accuracy of 84.41%. Out of the 15.59% of misclassified exemplars, 3.65% of those misclassifications were correcting warped anomalies using uniform scaling. This is understandable, as it is known that a special case of DTW (open-ended DTW) is essentially equivalent to uniform scaling.

The second most common misclassification was a Step anomaly remedied by applying some amount of downward linear trend. Oftentimes, if the step anomaly is a higher mean in the latter half of the anomalous subsequence, applying some downward linear trend will produce a similar effect as piecewise normalization. These accounted for 2.9% of the errors.

PUPAE can explain the vast majority of anomalies in the benchmark successfully. We show three examples in Figure 15.

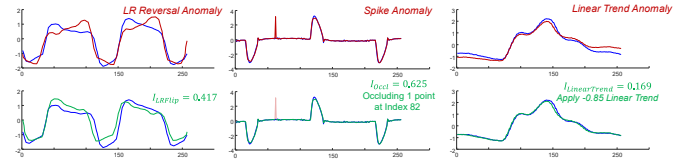

Figure 15: Three random examples of successful predictions on the generated benchmark dataset. The domains from left to right are insect gait, electrical demand fingerprinting, and PPG. *Top)* The nearest anomaly-free ED neighbor (blue) to the anomaly, and the anomaly (red). *Bottom)* The nearest anomaly-free ED neighbor (blue) to the anomaly, and the transformed anomaly according to $f_\theta^*$ (green), where $f_\theta^*$ is selected according to Equation 1.

For brevity, we relegate other examinations of misclassifications to [49]. Details about DAMP's success w.r.t. each anomaly class and domain are also housed at [49].

# 7 Conclusions and Future Work

We have shown a simple framework for explaining anomalies. We have shown both significant anecdotal evidence that it works and demonstrate its performance on a large-scale benchmark data based on an Amazon anomaly generator [17]. Beyond the direct utility of explaining anomalies, we believe our ideas can be used to improve the user experience of using any TSAD algorithm. One the most cited problems with deployments of TSAD algorithm is false alarm fatigue [37]. If it is noted that most false alarms have similar explanations, that observation could be exploited to improve the algorithms. We leave such explorations for future work.

Finally, we have created a new benchmark of 9,600 time series with ground truth annotations. We believe that that this dataset will help others to research the task-at-hand at [49].

SUPPLEMENTARY FILE

**Two More Case Studies:**

**A Particle Accelerator Case Study**

The **particle accelerator** dataset is a recently released dataset that contains "*anomalies in the high voltage converter modulator*" [30]. The dataset is interesting for several reasons. Normal behavior is polymorphic. Many of the anomalies are trivially obvious, but some are *very* subtle, and impossible to detect with the naked eye. In addition, the sampling rate is extraordinarily high. For example, this ⌐‾‾‾‾ normal pulse represents only 0.002 seconds, but is comprised of 4,500 datapoints. While the only papers to examine this archive so far use a variety of deep learning methods, we found that the simple idea outlined in IV.*A* to be at least competitive.

How can we explain this anomaly ⌐‾‾‾‾, which visually *seems* to be identical to the normal pulse above. Figure 16 offers a hint as to PUPAE's answer to this question.

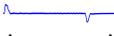

Figure 16: These two time series have a Euclidean distance of 63.0, but the distance plunges to 21.1 if we use DTW distance.

Given the large reduction in error for DTW, PUPAE states:

- Anomaly ⌐‾‾‾‾ would be like **DTL_Normal_13_3**, **except** for **warping of upto 6%**

Unfortunately, there is no ground truth for the causes of the anomalies, however the result here seems at least plausible.

**An Industrial Case Study**

We conclude with an end-to-end case study in anomaly *detection* and *explanation*. We investigated a dataset that reflects the behavior of a high-speed multilevel conveyor belt system [7], a short video of the system can be seen here [49]. The data has unusually good provenance, as the anomalies were induced by technicians physically interfering with the system in various ways. In Figure 17.*top* we show the OwBHLpower trace.

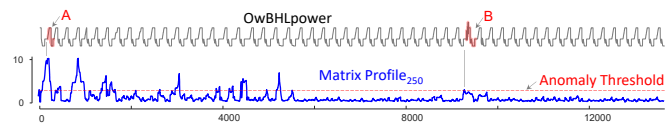

Figure 17: *top*). An excerpt from the test dataset OwBHLpower. *bottom*) The Matrix Profile of OwBHLpower. Selected sections that are above the anomaly threshold (red-dashed-lined) are highlighted.

We chose 250 as the subsequence length here (about 12 seconds of wall clock time), as visual inspection suggests it is approximately one cycle length. Using *only* training data (not shown) we computed the anomaly threshold of 4.1 by simply computing the Matrix Profile on the training data and finding the maximum value, plus one standard deviation. As shown in Figure 17.*bottom* we computed the Matrix Profile on the test data and used the anomaly threshold to separate the out anomalies. There are 12 anomalies discovered, 11 of which are true positives. In Figure 18 we show one of the most subtle anomalies, and one of the most obvious anomalies.

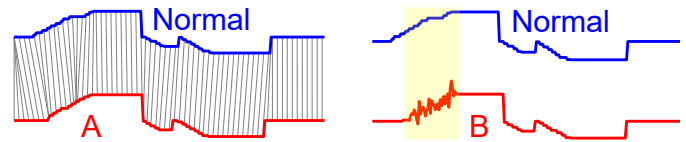

Figure 18: *Left*). Anomaly A is best explained by *warping*. *Right*) A Anomaly B is best explained by a *noise*.

The explanations created by PUPAE for these anomalies are:

- Anomaly **A** would be like **Train$_{4323}$**, **except** for **warping of upto 2.7 seconds**
- Anomaly **B** would be like **Train$_{4325}$**, **except** for **noise from 465.8 to 467.0 seconds**

The explanation for B is clearly true, and the explanation for A is consistent with the original author's explanation of how they induced some of the anomalies with a mechanical intervention that lead to the "shortening of cycles" [7].

Thus, at least in this domain, we can quickly set a single parameter, and then correctly discover, and plausibility explain time series anomalies.

**Miscellaneous Notes**

In the main paper we noted "Moreover, they are *trivial* anomalies in the sense, noted by Wu & Keogh [45].". To give an example: we can detect cutoffs of length five with this MATLAB snippet:

```
movmean(diff(xx)==0,5)==1.
```

Wu and Keogh have argued that these should *not* be used to evaluate TSAD.